\documentclass[journal]{IEEEtran}
\usepackage{widetext}
\usepackage{booktabs}
\usepackage{multirow}

\usepackage[normalem]{ulem}

\ifCLASSINFOpdf
  \usepackage[pdftex]{graphicx}
  \graphicspath{{../pdf/}{../jpeg/}}
  \graphicspath{{../figures/}}
  \DeclareGraphicsExtensions{.pdf,.jpeg,.png}
\else
  \usepackage[dvips]{graphicx}
  \graphicspath{{../eps/}}
  \DeclareGraphicsExtensions{.eps}
\fi
\usepackage{xcolor}

\usepackage{amsmath}
\usepackage{amsfonts}
\usepackage{bm}

\usepackage{subcaption}

\begin{document}
%
\title{Distribution prediction for image compression: \\ An experimental re-compressor for JPEG images \\
\vspace{0.5cm}
\large{\bf Huawei Technical Report \\}
\vspace{0.3cm}
\small{December, 2021}}

%
%
%
\author{
    \IEEEauthorblockN{Maxim Koroteev\IEEEauthorrefmark{1}, Yaroslav Borisov\IEEEauthorrefmark{1}, and Pavel Frolov\IEEEauthorrefmark{1}}
    \\\IEEEauthorblockA{\IEEEauthorrefmark{1} \small Huawei Russian Research Institue, Moscow, Russia
    \\}

\thanks{Moscow Research Center, Moscow, Russia. Cloud BU.  E-mail: koroteev.maxim@huawei.com}}

%
%

\markboth{Huawei Technical Report, December 2021}
{Koroteev \MakeLowercase{\textit{et al.}}:}
%



\maketitle

\begin{abstract}
We propose a new scheme to re-compress JPEG images in a lossless way. Using a JPEG image as an input the algorithm partially decodes the signal to obtain quantized DCT coefficients and then re-compress them in a more effective way.
\end{abstract}


%
\IEEEpeerreviewmaketitle

\section{Introduction}
%
%
%
%

\IEEEPARstart{I}{n} recent years with emerging multiple various data storage and cloud services the problem of data compression became vital again. One of the key and most expensive factors for organizing cloud services is disk space. On the other hand, the most part of the data currently located in the cloud is media data. This data is normally already compressed in some more or less efficicent way but still it is very desirable to improve its compression further to save the disk space in the cloud on a larger scale.

There may be multiple approaches to this problem varying in scale and compelxity and related to building models of the data of various types. In this paper we focus on a more traditional approach of compressing or rather re-compressing media data, namely, we try to propose some more efficient algorithms for better compression of individual compressed image files. We also focus on jpeg files here as among media data in the cloud video and jpeg compressed files represent a significant fraction.


\section{An approach to re-compress jpeg images}
\label{sec:approach}

When uploading media data into the cloud it is usually assumed by the user that the uploaded data can be retrieved back exactly in the same form as it was uploaded. In other words, no one expects the cloud algorithm would distort the data handled to it. 

Therefore, when working on re-compression of media data in this assumption, the only possible apporach can be lossless compression. It is well known that the main source of losses in compression of media data is quantization; when quantized, media data is normally compressed losslessly. In the currently used approaches to re-compression, which we follow as well, we can try to improve the algorithms following quantization, i.e., to re-compress quantized DCT coefficients. Thus it is assumed below that a re-compression algorithm extracts quantized DCT coefficients out of the compressed data stream and transmits them to a core re-compressor which should compress them in a better way providing an improved compression gain.

\section{Some remarks on the statistics of DCT coefficients}
\label{sec:statistics}

\begin{figure}
\centering
\includegraphics[width=1.01\linewidth]{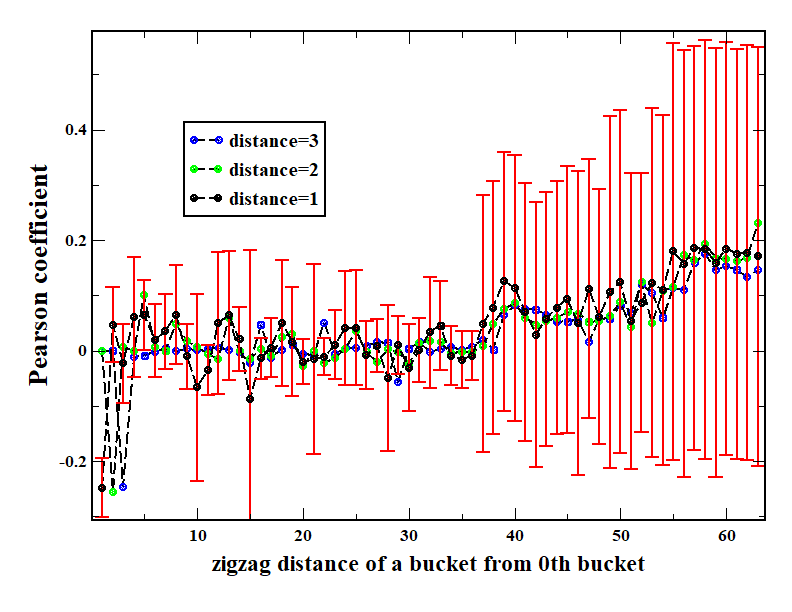}
\caption{For each image of a dataset consisting of $18$ widely spread jpeg images (see the table at the end of the paper) the Pearson correlation coefficient has been computed between DCT coefficients located at different position of $8\times 8$ blocks. All distances are measured in terms of zigzag order. Note that the error bars are shown only for one experiment; for other they are similar. The only statistically significant deviation of the correlation coeffcient from zero is observed for measuring correlations with top left position (DC coefficient position). Even in this case the correlations are extremely weak ($\sim 0.2$).}
\label{fig:bucket_correlations}
\end{figure}

JPEG algorithm deals with $8\times 8$ blocks of DCT coefficients, so each block contains $64$ coefficients. It is well known that the application of the discrete cosine transform results in a certain decorrelation between the adjacent pixels of the block, so that the final DCT coefficients are sufficiently independent in terms of {\it linear correlation}. To illustrate this we show the correlation coefficient measured between the neighbouring positions of $8\times 8$ blocks for a dataset of jpeg images (fig. \ref{fig:bucket_correlations}).
It is seen that the linear correlations are insignificant for all locations of DCT coefficients except when measuring correlations between the DC coefficients and some AC coefficients closest in terms of the zigzag distance. However, it is quite natural even in the latter case the correlation magnitude remains at the level $\sim 0.2$, which is too small for building a realistic model out of it. Some increase of standard deviations observed in the right part of fig. \ref{fig:bucket_correlations} is accounted for by the significant fraction of zero DCTs at positions of the block remote from the top left corner (the position of DC coefficient). These observations remain sufficiently stable for various images and indicate that efforts to build {\it linear} prediction models based on information from the adjacent positions in the block are doubtful. Let us reiterate, more sophisticated models which would take into account higher order statistics can be of use; they require further analysis and will not be considered in this paper.

\begin{figure}
\centering
\includegraphics[width=1.01\linewidth]{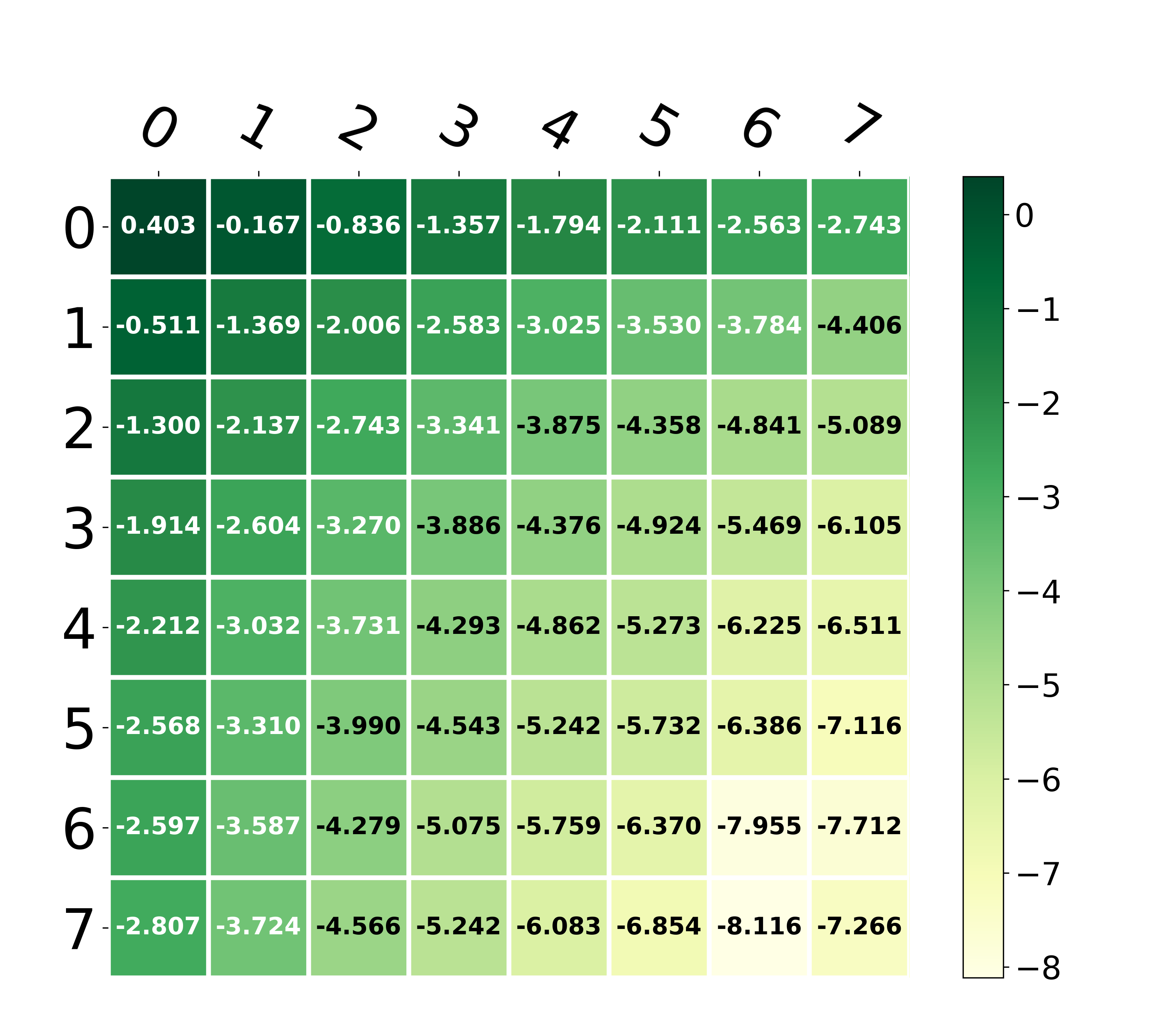}
\caption{Distribution of standard deviations for DCT coefficients computed for various positions inside $8\times 8$ block of an image. Note the log scale to clearly indicate the trend. All bllocks of an image have been collected and stds for quantized DCT coefficients have been computed for each position in the block. The logarithm of the magnitude of std is shown for each position.}
\label{fig:std_distrib}
\end{figure}

Next, DCT transform typically results in forming a certain pattern in statistics of DCT coefficients located in different parts of the block. To illustrate this property we plot the behavior of the second moment of the distribution (standard deviation) of DCT coefficients at each position of an image (fig. \ref{fig:std_distrib})\footnote{such a computation is merely suggestive; the standard assumption about DCT coefficients located at different positions of the block, confirmed by computations, is that their distribution is non-stationary so in general DCT coefficients at each position of a block and {\it for each block} are taken from {\it different distributions}.}. Based on this well-known observation we consider the compression of DCT coeffcients separately for each position in the block which corresponds to separate encoding of the coefficients corresponding different frequencies. To shorten the explanations we will refer, rather informally, to this procedure as {\it bucketing}; so we imply below that each of $64$ positions in $8\times 8$ block corresponds to one bucket and thus we consider the compression of $64$ buckets. 
Note that implementation of this approach usually does not require creating separate physical buffers in memory for buckets; instead, it is sufficient to use appropriate pointer shifts in the array of all DCT coeffcients. 

\section{Probability model}

\begin{figure}
\centering
\includegraphics[width=1.01\linewidth]{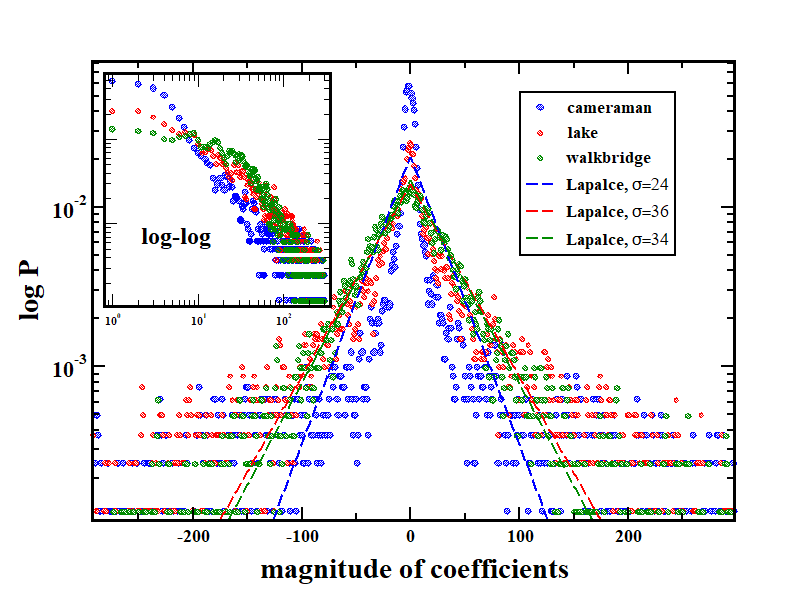}
\caption{Distribution of $\Delta$DC coefficients (i.e., first forward differences between DC coefficinets for an image) for several images from the dataset. Note semi-log scale for more convenient comparison with exponential distributions. For each empirical distribution we also plot the Lapalce distribution for $\mu=0$ and std shown in the legend. These stds correspond to those measured in empirical distributions. Note also the log-log inset which demonstrates longer tails for the empirical distributions.}
\label{fig:delta_dc_distrib}
\end{figure}

After separating all DCT coefficients of an image into buckets and applying delta coding to the first bucket containing DC coefficients, we would like to predict the probability of each coefficient in each bucket to use these probabilities in a multisymbol arithmetic coder (MSAC). From the point of view of adaptive probability prediction this aprroach implies forming certain contexts for each position of DCT coefficients. As a probability model for all buckets the Laplace distribution is usually used which appears to have correspondence to actual distributions of DCT coefficients and in the same time sufficiently simple for computations. However, it is clear that when dealing with non-stationary distributions it is difficult to confirm that the distributions inside buckets remain laplacian or at least keep their parameters fixed (usually they do not). Moreover if we look at the distributions of DC coefficients, the deviation from Lapalce becomes evident (fig. \ref{fig:delta_dc_distrib})

\begin{figure}
\centering
\includegraphics[width=1.01\linewidth]{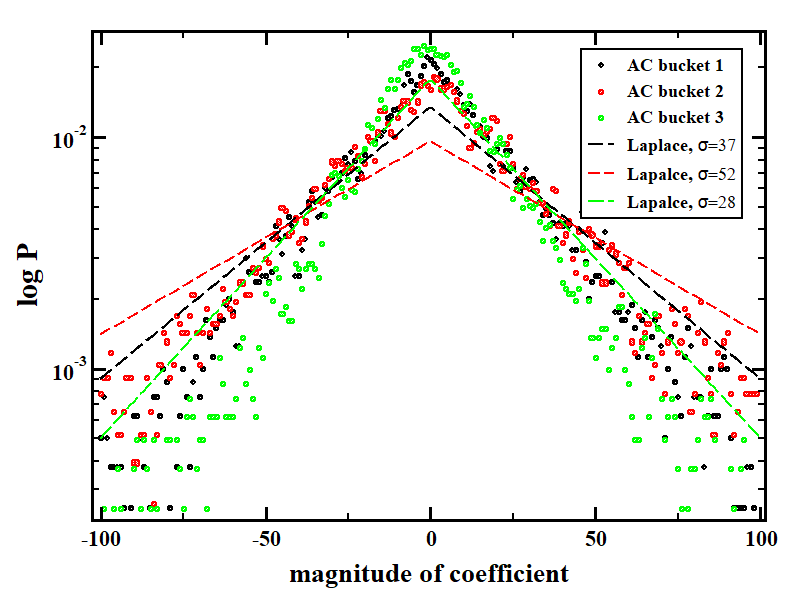}
\caption{Empirical normalized distribution of AC coefficients in the first three buckets for walkbridge image. Note semi-log scale for more convenient comparison with exponential distributions. For comparison three Lapalce pdfs are shown, with $\sigma$ estimated from the data.}
\label{fig:ac_distrib}
\end{figure}

In fig. \ref{fig:ac_distrib} we show the distributions of AC coefficients for an image compared to the pdfs for Lapalce distributions.
It is seen that various deviations from the Lapalce distribution may occur both in the region of small amplitudes of the coefficients and in tail regions. However, there is an essential difference between figs. \ref{fig:delta_dc_distrib} and \ref{fig:ac_distrib}. For DC buckets the distributions systematically demonstrate longer tails close rather to power-law behavior (see inset on the fig. \ref{fig:delta_dc_distrib}), while the distributions for AC keep the laplacian form: even though the comparison with Laplace distributions shows some discrepancy, the distributions remain quite straight in the semi-log scale indicating that the approximation with Lapalce distribution proves to be adequate for AC coefficients and it is just the question of better estimate for the standard deviation of this distribution which matters; this is valid for majority of observed cases. On the other hand, for DC coefficients an approximation with a distribution with longer tails and sharper peak near zero (e.g. generalized gaussian) may be a better choice. We will discuss this at some other point but for a moment we will assume Lapalce distribution for DC coefficients as well.

We apply the Lapalce distribution to the probability computation adaptively, i.e., for each of $64$ buckets of the block we assume that the probability density for an underlying random variable $x$ has the form
\begin{equation}
\label{laplace_pdf}
p(x, \mu, \sigma) = \frac{1}{2\sigma}e^{-\frac{\left|x-\mu\right|}{\sigma}}.
\end{equation}
This random variable will approximate the distribution for the original alphabet of DCT coefficients, for which we additionally take $\mu=0$. They are also assumed conditionally independent given variances. Then the distribution function will be
\begin{equation}
\label{distr_function}
F(x,\sigma) = \begin{cases}
\frac{1}{2}e^{x/\sigma}, & x<0 \\
1 - \frac{1}{2}e^{-x/\sigma}, & x\ge 0.
\end{cases}
\end{equation}
Now, if $X$ is a discrete alphabet $\ldots -3, -2, -1, 0, 1, 2, 3,\ldots$, then the probability\footnote{Generally speaking this is a conditional probability as it depends on the distribution parameter for the previous coefficients in the bucket.} for each symbol would be approximated by means of (\ref{distr_function}) as
\begin{equation}
\label{probability}
P(X_{i}, \sigma) = F(X_{i}+0.5, \sigma) - F(X_{i}-0.5, \sigma).
\end{equation}
The last formula was used for actual probability computations in the implementation. To make all necessary computations faster probability tables corresponding to various $X$ and $\sigma$ have been computed using (\ref{laplace_pdf}), (\ref{distr_function}) and (\ref{probability}).

\section{Prediction}

To compute the probability we have an additional parameter $\sigma$, the second moment of the Lapalce distribution. As the sequence of DCT coefficents can not be considered as stationary, but as we still would like to exploit the redundancy for each bucket, we need to model variances $sigma$ in each subband. For that end we can try GARCH-like model approach to predict the second moment of the time series\footnote{see also the next section for additional discussion}. Then this parameter can be predicted for each bucket using the simplest exponential smoothing approximation for a time series $x$:
\begin{equation}
\label{expon_smoothing}
\sigma_{i} = \beta\sigma_{i-1} + \alpha |x_{i-1}|.
\end{equation}
This is a purely empirical model and can be justified in several ways, e.g., by the following obvious argument. If a random variable $X$ is distributed according to (\ref{laplace_pdf}) and if $n$ observable values of $X$ are $x_{1}, x_{2},\ldots, x_{n}$, then the logarithmic likelihood function has the form
$$
\log L = \log\frac{1}{(2\sigma)^{n}} - \frac{1}{\sigma}\sum_{i}x_{i}.
$$
Differentiate the last expression wrt $\sigma$ to find where the maximum of $L$ is attained. We have 
$$
\frac{\partial\log L}{\partial\sigma} = -\frac{n}{\sigma} + \frac{1}{\sigma^{2}}\sum_{i}|x_{i}| = 0.
$$
It follows that the maximum likelihood estimate (MLE) of $\sigma$, which provides the maximization of an event when $x_{1}, x_{2},\ldots, x_{n}$ occur yields
$$
\sigma = \frac{1}{n}\sum_{i}|x_{i}|,
$$
which for the case $n=1$ takes on an especially simple form: $\sigma = |x|$\footnote{Another method to justify the exponential smoothing model would be just to try to write down the distribution for a random variable $y=|x|$ and calculate its expectation. It is not difficult to show that $\langle y\rangle=\langle |x|\rangle=\sigma$ if $X$ is distributed according to Laplace. Loosely speaking this allows to interpret (\ref{expon_smoothing}) as a kind of "mean field" approach.}. Thus the best estimate of $\sigma$ in case of Laplace distribution is just the absolute value of the observable $x$. Therefore the simplest possible model for $\sigma$ would have the form $\sigma\sim |x|$ but to take into account a weak dependency on the previous state we introduce a minimum generalization which immediately results in (\ref{expon_smoothing}). It is necessary to stress that the above argumentation implies maximization of the posterior probability and thus implies the prior not depending on $\sigma$, e.g., a uniform one.

Note, that in GARCH model the prediction is usually done for the variance of the random variable, so (\ref{expon_smoothing}) can be written as
\begin{equation}
\label{expon_smoothing_variance}
\sigma^{2}_{i} = \tilde{\beta}\sigma^{2}_{i-1} + \tilde{\alpha} |x_{i-1}|^{2}.
\end{equation}
The justification of this model follows from the previous consideration. In practice, it seems there is no way to decide which model to prefer: it should be determined in numerical experiments.

The models like  (\ref{expon_smoothing}) or (\ref{expon_smoothing_variance}) are extremly generic and no wonder they already appeared from time to time in various compression related applications (see e.g. \cite{zhang_icip_2014}). Interestingly, they did not seem to receive much attention.

\section{Connection of the prediction model to a random process}

The way formula (\ref{expon_smoothing}) was introduced in the previous section was purely empirical and one would be interested in a more strict interpretation or at least in a comparison with some mathematically more clear construction. The easiest way to do this for a non-stationary signal is to compare it to some elementary random process.

Let us assume we construct a random process in the following way.
$$
\sigma_{1} = \beta\left|Z_{0}\right|,
$$
$$
\sigma_{2} = \alpha\sigma_{1} + \beta\left|Z_{1}\right|,
$$
$$
\ldots
$$
$$
\sigma_{k} = \alpha\sigma_{k-1} + \beta\left|Z_{k-1}\right|,
$$
Here $Z_{i}$ are random variables having the Lapalce distribution with expectation $EZ_{i}=0$ and variance $DZ_{i}=s^{2}$. From the above formulas the general term $\sigma_{k}$ can be easily re-written as
\begin{equation}
\sigma_{k} = \beta\left(\alpha^{k-1}\left|Z_{0}\right| + \alpha^{k-2}\left|Z_{1}\right|+\ldots +\left|Z_{k-1}\right|\right).
\label{approx_sigma}
\end{equation}
Noting that from $EZ_{i}=0$, $DZ_{i}=s^{2}$ it follows $E\left|Z_{i}\right|=s$, $D\left|Z_{i}\right|=s^{2}$ and then the last formula yields for the expectation and variance of $\sigma_{k}$
$$
E\sigma_{k} = \beta s\frac{1+\alpha^{k-1}}{2}k, \; D\sigma_{k} = (\beta s)^{2}\frac{1+\alpha^{2k-2}}{2}k.
$$
Thus, as in the case of a simplest random walk the process represents a random walk in the neighbourhood of a line with the slope $\beta s (1+\alpha^{k-1})/2$ with the increasing deviations proportional to $\beta s \sqrt{(1+\alpha^{2k-2})/2}\sqrt{k}$. Let us assume $\left|\alpha\right|<1$, then for $k\gg 1$ the term with $\alpha$ becomes negligible and the process behaves as
$$
E\sigma_{k}\sim \beta sk, \; D\sigma_{k}\sim (\beta s)^{2}k.
$$
So the basic behavior is determined by the standard deviation of random variables $Z_{i}$. The magnitude of $\alpha$ controls the influence of adding $\left| Z_{i}\right|$ into the $\sigma_{k}$. In our applications we deal with the situation when all $Z_{i}$ 1) have different standard deviations; 2) these standard deviations are unknown. The former problem can be simplified by choosing smaller $\alpha$, which results in only the closest terms affecting $\sigma_{k}$ at the step $k$ in (\ref{approx_sigma}); for the latter we have to provide an estimate for $s$. This is what we discussed in the previous section where it was demonstrated $s\sim \left|Z_{i}\right|$, which results in the estimate $E\sigma_{i}\sim\beta\left|Z_{i}\right|$.

\section{Adaptation of the prediction model to the encoding scheme}

It is not difficult to understand (and check experimentally) that the direct use of the models (\ref{expon_smoothing}), (\ref{expon_smoothing_variance}) has a restricted efficiency. The main reason for that is that the models imply some 1D time series in which the connection strength between observable values decreases in some proportion of the linear distance between these observables. But this is not the case for the data organized in a 2D structure, like images. The dependences here are also 2D and any 1D representation of the coefficients in the buckets can not be described in terms of the linear distance. It can be expressed another way by saying that contexts for a particular coefficient in a bucket turn out to be more complicated and talking about some context of previous coefficients, the word previous should be understood not in terms of some linear distance. To take this into account we make some modifications for the basic exponential smoothing model.

We start from formula (\ref{expon_smoothing}) and apply it to each bucket as\footnote{the following discussion can be literally repeated if instead of using the standard deviation $\sigma$ one uses variance $\sigma^{2}$; it is a matter of experiments with the model as we pointed out at the end of the previous section.}
\begin{equation}
\label{predictor}
\sigma^{k}_{i} = (1 - \alpha)\sigma^{k}_{i-1} + \alpha |DCT^{k}_{i-1}|.
\end{equation}
In this formula $k$ is the number of the bucket, i.e., $k=1,2,\ldots 64$; $\sigma^{k}_{i}$ is the magnitude of the second moment for bucket $k$ for block $i$\footnote{The counting of the blocks should not necesserily follow the raster order; one can define more complicated contexts for $\sigma^{k}_{i}$.}; $|DCT^{k}_{i-1}|$ is the absolute value for the DCT coefficient in bucket $k$ of the block $i-1$ and $\alpha$ is an empirical parameter. The formula for variances will have the same form; we deal with (\ref{predictor}) for the sake of simplicity.

Now to take into account 2D dependences we predict $\sigma$ in three directions: horizontal, vertical, and diagonal, the approach very well known in compression algorithms. Thus, we compute $\sigma$ for each bucket three times which corresponds to consideration of three 1D time series (horizontal, vertical, and diagonal), the length of which is confined by the sizes of an image. This yields three predicted $\sigma$: $\sigma^{k}_{ih}, \sigma^{k}_{iv}, \sigma^{k}_{id}$; the resulting standard deviation is computed
as
$$
\sigma^{k}_{i} = A\sigma^{k}_{ih} + A\sigma^{k}_{iv} + B\sigma^{k}_{id},
$$
where wheight $A$ for the horizontal and vertical components are taken to be equal and normally $B<A$\footnote{these parameters are subject to adaptation and the releation between them can be changed; we just provide our empirical observations for the best set of the parameters.}. 

The prediction of sigma can be further improved by taking into account a (weak) dependency existing between various buckets. This can be presented in different forms; to give an idea how this can be done we can collect sufficiently large statistics of linear dependencies between the pairs of buckets just by measuring Pearson linear correlation coeffcient and then for each bucket to rank other buckets wrt. this coefficient. We then apply the exponential smoothing model as follows
$$
\sigma^{k}_{i} = \gamma\sigma^{k}_{i-1} + (1 - \gamma)|DCT^{m}_{i-1}|,
$$
where $m$ is the index of the most dependent (in terms of linear correlation) bucket.

When the second moment is estimated using the above model, the probabilities of the DCT coefficients can be computed using pre-computed tables. After that the symbols and their probabilites can be transmitted to the multisymbol arithmetic coder (MSAC). 

\section{Application of RLRG encoder in the encoding scheme}

It was noted that for buckets located in the right bottom corner of the block the lengths of runs of zeros become significant. Therefore we found slightly more effective to use for these buckets another method, the adaptive run-length Rice-Golomb encoder (RLRG). This encoder does not use the moment estimations provided above but exploits its own small set of parameters which can be adaptet for a particular task. The detailed discussion of this algorithm is provided in \cite{Malvar}.

This algorithm, as it name indicates, is referred to run-length algorithms and consequently can be effective when encoding long runs of repetitive symbols. In the context of DCT coefficient compression it is applied to our buckets when we have long runs of zeros. Another important feature of this algorithm is its lower complexity compared to the arithmetic coder. 
We found it provides better compression ratio for buckets with smaller $\sigma$. 

In terms of practical computations we measured the number of zeros for each bucket for all images in the data set and empirically chose the boundary in terms of the bucket counter. For all buckets with significant number of zeros ($>75\%$) RLRG encoder was used instead of MSAC. After analyzing sufficiently large number of images one can determine a threshold for the number of buckets encoded with RLRG.

\section{Remarks on implementation}
\label{sec:implementation}
\begin{figure}
\centering
\includegraphics[width=0.45\linewidth]{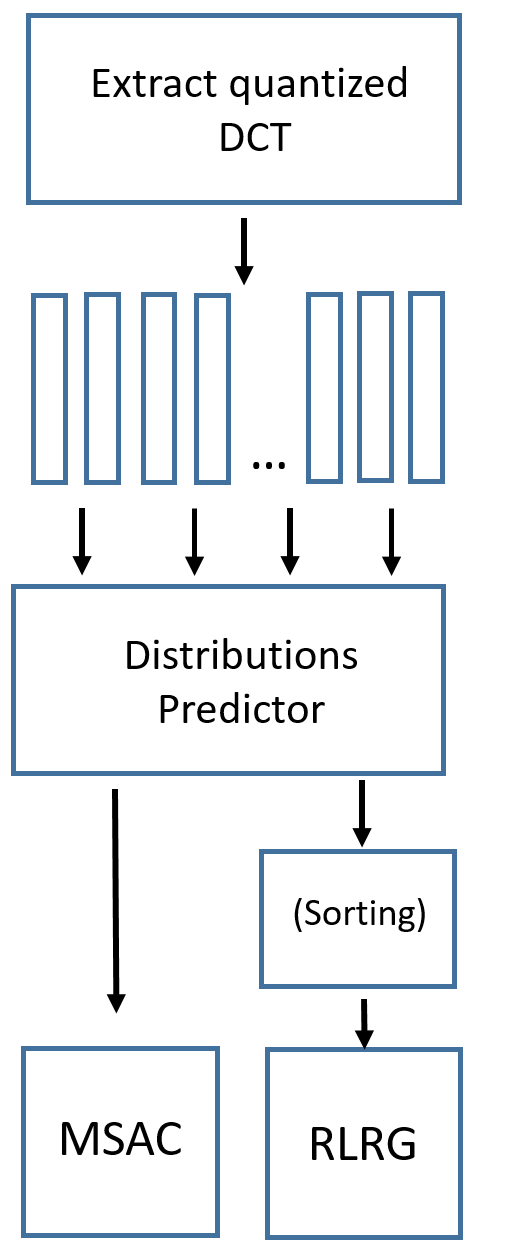}
\caption{Generic layout of the re-compression algorithm. The blocks are described in the main text. Block 'sorting' is indicated as optional and applied to the bukets compressed with RLRG encoder.}
\label{fig:diagram}
\end{figure}

The general layout of our approach to re-compression is presented in fig. \ref{fig:diagram}.
The algorithm takes on as an input a compressed jpeg stream and decodes quantized DCT coeffcients losslessly; we used a modified ffjpeg software for that. Then we apply our prediction algorithm and finally two entropy coders: MSAC and RLRG, which are used as explained in the previous section. For MSAC case probability tables have been pre-computed based on analytic formulas for the Laplace distribution. Note, that the coefficients in buckets compressed with RLRG can be additionally sorted in the decreasing order of the coefficients in the previous bucket (so this order is available in the decoder); this can help improve compression gain in some situations but this improvement is moderate.

\section{Re-compression results}

As a test set we used a range of various jpeg images both color and grey scale, most of them being used in many image compression experiments. The compression results obtained with the proposed method are shown in table \ref{tab:results} and include both luma and chroma data.
\begin{table}[htbp]
	\centering
	\caption{Re-compression results. The table shows the results of compression gain measurements using the proposed algorithm. The second column indicates the size of the original jpeg file in bytes and the third column -- the size of the re-compressed file. Averaged gain achieved on this data set was measured on the level $14,5\%$}
\resizebox{\columnwidth}{!}{%
	\begin{tabular}{lrrl}
		\toprule
		\multicolumn{1}{c}{image name} & \multicolumn{1}{c}{original size}  & \multicolumn{1}{c}{re-compressed size} & \multicolumn{1}{c}{gain, \%}\\ 
		\midrule
		cameraman   & 50483  & 42914 & 14.99\\
		
		cat               & 105552 & 89657 & 15.06 \\
		
		house           & 36002   & 29871  & 17.03 \\
		
		jetplane       &  61553  & 52789   & 14.21 \\
		
		lake            &  87217  & 74056   &  15.09 \\
		
		lena-color-256  & 29309 & 25426 & 13.25 \\

		lena-color-512  & 105733 & 89768 & 15.1 \\

		lena-grey-256  & 21133   & 18428  & 12.8 \\

		lena-grey-512  & 64762  &  54926 & 15.19 \\

		livingroom  & 82420     &   70353 & 14.64 \\

		mandril-color & 247196 & 214342 & 13.29 \\

		mandril-grey & 88031   & 72627  & 17.5 \\

		peppers-color & 143447 & 125098 & 12.79 \\

		peppers-grey & 78743   & 68092  & 13.53 \\

		pirate   & 80393       & 70154      & 12.74 \\

		walkbridge  & 109050  & 93211   & 14.52 \\

		woman-blonde  & 80295  & 69025  & 14.04 \\

		woman-darkhair & 45172  & 38102 & 15.65 \\
		\bottomrule
	\end{tabular}%
}
	\label{tab:results}%
	
\end{table}%

On the other hand, in industrial applications the use of two encoding algorithms seems to be not an optimal solution in terms of simplicity for implementation. However, it is not a complicated task to remove one of the encoders under consideration and apply another one for all the buckets. We carried out this computation and provide results in table  \ref{tab:MSAC} for the case when MSAC is used for all buckets.

\begin{table}[htbp]
	\centering
	\caption{Re-compression results. MSAC applied for all buckets.}
\resizebox{\columnwidth}{!}{%
	\begin{tabular}{lrrl}
		\toprule
		\multicolumn{1}{c}{image name} & \multicolumn{1}{c}{original size}  & \multicolumn{1}{c}{re-compressed size} & \multicolumn{1}{c}{gain, \%}\\ 
		\midrule
		cameraman   & 50483  & 42759 & 15.3\\
		
		cat               & 105552 & 89856 & 14.87 \\
		
		house           & 36002   & 29951  & 16.81 \\
		
		jetplane       &  61553  & 52946   & 13.96 \\
		
		lake            &  87217  & 74682   &  14.37 \\
		
		lena-color-256  & 29309 & 25539 & 12.86 \\

		lena-color-512  & 105733 & 91860 & 13.12 \\

		lena-grey-256  & 21133   & 18429  & 12.8 \\

		lena-grey-512  & 64762  &  55856 & 13.75 \\

		livingroom  & 82420     &   71162 & 13.66 \\

		mandril-color & 247196 & 215637 & 12.77 \\

		mandril-grey & 88031   & 72481  & 17.66 \\

		peppers-color & 143447 & 136344 & 4.95 \\

		peppers-grey & 78743   & 69546  & 11.68 \\

		pirate   & 80393       & 70617      & 12.16 \\

		walkbridge  & 109050  & 93152   & 14.58 \\

		woman-blonde  & 80295  & 69707  & 13.19 \\

		woman-darkhair & 45172  & 38586 & 14.58 \\
		\bottomrule
	\end{tabular}%
}
	\label{tab:MSAC}%
	
\end{table}%

Excluding significant loss occurring on {\it peppers-color} image other images demonstrate slightly worse resutls than the original scheme. Overall gain loss, excluding peppers-color, is $0.5\%$, which can be thought of as insignificant even though the gain loss is observed on the majority of test images and consequently is systematic. Therefore for purposes of simpler implementation one MSAC encoder can be used for all buckets and potentially it seems it should be possible to outperform the combination of MSAC and RLRG by the single MSAC by improving the contexts; this, if possible, will be presented elswhere.

\section{Discussion}

Any compression scheme is not ideal. The proposed solution for re-compression of images has its pros and cons. On the one hand, the model for the distribution parameter prediction is simple, being based on explicit prediction formulas, not being buried under a bunch of contexts, and easy to implement. Moreover, the use of pre-computed tables enables to avoid time loss on their computation in the process of encoding.

One of the goals of this paper was to provide a single algorithm solution for purposes of re-compression. But in the same time we currently can notice an opposite tendency: some modern approaches to media data compression mix up multiple methods, try to include multiple context models etc. It would be too naive to try to compete with complex methods. Among the recent achievements in the field of image re-compression the most noticeable is prrobably {\it brunsli}\cite{brunsli} which represent a pretty nice combination of good ideas, some of which are similar to those used in our approach\footnote{Our test show that our re-compressor implementation is inferior to brunsli in terms of compression gain by $2\%$ on average. Yet we have to note that brunsli utilizes additional procedures for gain improvement related to histogram representation.}. Another purpose for implementing the approach with two entropy encoders was to demonstrate a good potential for RLRG algorithm compared to even such a powerful method as MSAC. The former algorithm looks elegant and somewhat underestimated. In the same time it does not require any probability computation and builds the context for encoding directly using the small set of adaptation parameters, providing lower complexity, as we already pointed out. However, the recent advances in the field of data compression show (quite naturally) that various sophisticated algorithms turn out to be competitive with each other in application to various data sources, i.e., many methods become data dependent\footnote{or overfitted}. This is certainly a sign of a some saturation achieved with application of classical compression methods. Nevertheless, the appearance of another compression scheme may be not without use, especially taking into account the fact this approach demonstrated a significant redundancy reduction over the standard algorithm.


%



\section*{Acknowledgment}

The authors are grateful to their colleagues in Huawei Algorithm Innovation Lab for discussions. MK is also grateful to Phil Chou for multiple stimulating discussinos in previous years which were partially embodied in this paper.
\ifCLASSOPTIONcaptionsoff
  \newpage
\fi



\bibliographystyle{IEEEtran}
\bibliography{bibtex/bib/IEEEabrv,bibtex/bib/IEEEexample}
%




%




\end{document}